\documentclass[conference]{IEEEtran}
\IEEEoverridecommandlockouts
\usepackage{cite}
\usepackage{amsmath,amssymb,amsfonts}
\usepackage{algorithmic}
\usepackage{graphicx}
\usepackage{textcomp}
\usepackage{xcolor}
\usepackage{graphicx}
\usepackage{amsmath}
\usepackage{amssymb}
\usepackage{booktabs}

\usepackage{amsmath,amsfonts}
\usepackage{array}
\usepackage{textcomp}
\usepackage{stfloats}
\usepackage{url}
\usepackage{bm}
\usepackage{color}
\usepackage{multirow}
\usepackage{booktabs}
\usepackage{verbatim}
\usepackage{graphicx}
\usepackage{subcaption}
\usepackage{ulem}
\def\BibTeX{{\rm B\kern-.05em{\sc i\kern-.025em b}\kern-.08em
    T\kern-.1667em\lower.7ex\hbox{E}\kern-.125emX}}
\begin{document}

\title{Patch Knowledge Transfer for Efficient AI-Generated Image Quality Assessment}

\author{
  \IEEEauthorblockN{
    Jiquan Yuan
  }
  \IEEEauthorblockA{ School of Software and Microelectronics, Peking University, Beijing, China}
  \IEEEauthorblockA{jiquany2023@163.com}
}


\maketitle

\begin{abstract}
With the rapid advancement of image generation technologies, perceptual quality assessment of AI-generated images has emerged as a crucial research direction in computer vision. The core challenge of this task lies in achieving efficient quality assessment for massive generated images. Current mainstream approaches exhibit two key limitations: 1) Methods employing complex feature extraction strategies, while improving performance, incur prohibitive computational costs that hinder real-time inference; 2) Simple image scaling-based solutions, despite their computational efficiency, demonstrate significantly inferior assessment accuracy. To address this critical issue, we propose Patch Knowledge Transfer (PKT), a knowledge distillation-based optimization framework that achieves synergistic optimization of visual representation capability and inference efficiency through an innovative multi-level knowledge transfer mechanism. Specifically, we design a dual-model architecture: a teacher model with local-global hybrid processing provides high-quality supervision signals, while a student model relying solely on global processing efficiently inherits the teacher's representation capacity through multi-level supervision. Extensive experiments conducted on 4 AIGIQA databases demonstrate that the PKT framework enables the student model to maintain performance comparable to the teacher while reducing computational costs by 67.7\%. Furthermore, compared to existing methods, our approach achieves a superior balance between model efficiency and assessment accuracy.
\end{abstract}


\section{Introduction}
With the rapid advancement of image generation technology, the number of AI-generated images (AIGIs) has surged, and they are now widely used across various fields. However, the quality of these AIGIs varies greatly, and low-quality AIGIs severely impair user experience. Therefore, efficiently filtering high-quality images from large-scale AIGIs is crucial.

To address this, the task of AI-generated image quality assessment (AIGIQA) has gained increasing attention. Most existing methods \cite{musiq,MANIQA,IP-IQA,yuan2023pkui2iqa,yuan2023pscr,IPCE_2024_CVPR,Yu_2024_CVPR,MOE-AGIQA, AMFF,AlignIQA,TSP} process image inputs in two primary ways: 1) global processing, which extracts features directly from the whole image—computationally efficient but prone to loss of fine-grained details; and 2) local–global hybrid processing, which aggregates features from multiple local patches and the global image—more accurate but computationally expensive and unsuitable for large-scale real-time assessment. This accuracy–efficiency trade-off becomes particularly pronounced for high-resolution images. Thus, we raise the key research question:  \textbf{\textit{Is there an innovative approach that can effectively explore and integrate multi-level visual representations (from local details to global semantics) while maintaining high computational efficiency?} }

\begin{figure}[!t]
\centering
\subfloat[PKT]{\includegraphics[width=0.49\textwidth]{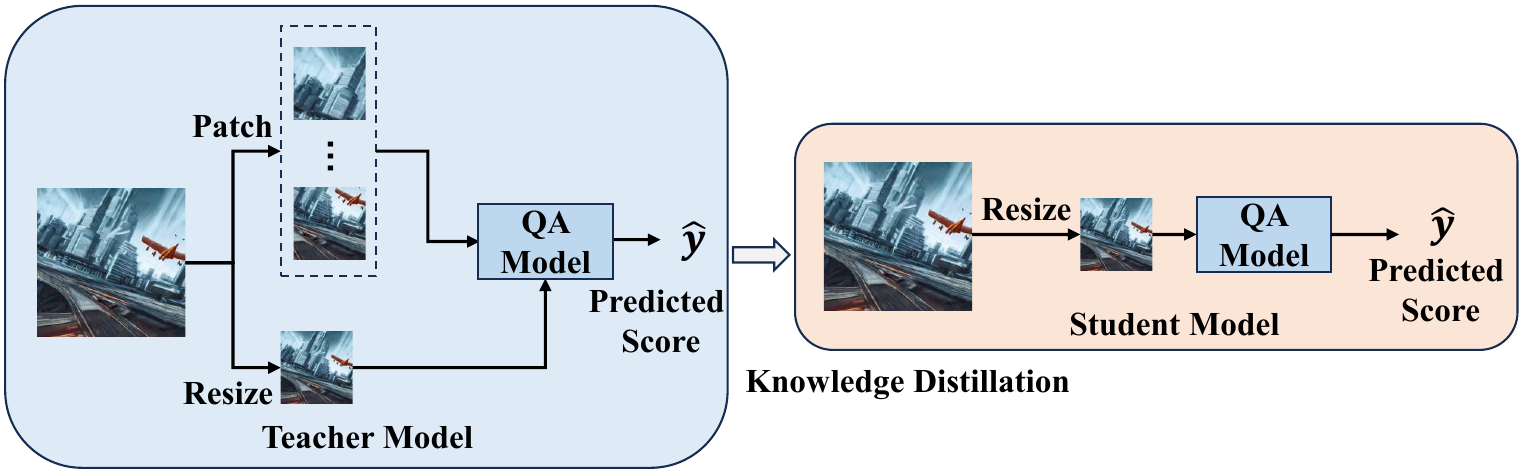}%
\label{KD}}
\hfil
\subfloat[SRCC vs. FLOPs]{\includegraphics[width=0.48\textwidth]{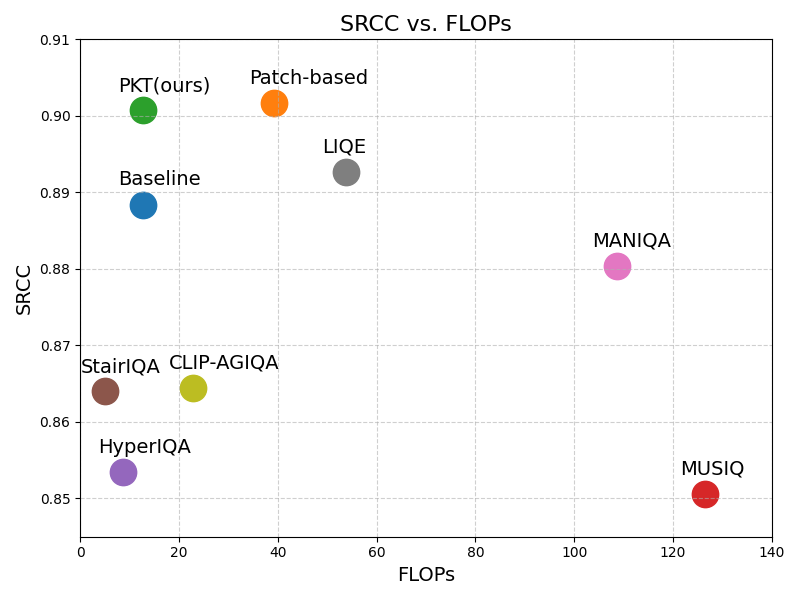}%
\label{SRCC}}
\caption{(a) Illustration of our proposed PKT, which employs a local-global hybrid teacher model to guide a global-only student model. (b) Comparisons with multiple patch-based teacher, baseline, and several existing IQA methods on AGIQA-1K database in terms of SRCC and FLOPs. }
\label{fig1}
\end{figure}

To tackle this challenge, we propose Patch Knowledge Transfer (PKT), a knowledge distillation-based framework that unifies comprehensive visual processing with efficient inference. As shown in Fig. \ref{KD}, we design a dual-branch architecture: the teacher model employs local–global hybrid processing to build comprehensive quality priors through multi-scale feature extraction; the student model retains only the global processing branch but learns the teacher’s multi-level representational capability via feature distillation. This allows the student model to maintain efficient single-image forward computation while inheriting the teacher’s rich quality discrimination knowledge, ultimately achieving an optimal balance between accuracy and efficiency (Fig. \ref{fig1}).

To ensure the student fully assimilates the teacher’s representations, we design a multi-level knowledge transfer scheme with two core supervision mechanisms: feature-level knowledge transfer, which aligns features at the last encoder layer via a feature  alignment loss, and output-level knowledge distillation, which minimizes the KL divergence between the two models’ prediction distributions. This hierarchical strategy transfers not only explicit discriminative knowledge but also the teacher’s internal feature extraction paradigm.

Extensive experiments on four mainstream AIGIQA databases demonstrate that PKT achieves a better trade-off between efficiency and performance, including AGIQA-1K \cite{AGIQA1K}, AGIQA-3K \cite{AGIQA-3K}, AIGCIQA2023 \cite{aigciqa2023}, and PKU-AIGIQA-4K \cite{PKUAIGIQA4K}. On AGIQA-1K, compared to the local–global hybrid teacher model, PKT reduces computational cost (FLOPs) by 67.7\% while maintaining comparable performance; compared to the global-only baseline, it improves SRCC and PLCC by 0.0098 and 0.0091, respectively, under the same computational budget. Notably, the student model even surpasses the teacher on certain metrics, indicating effective knowledge transfer and performance enhancement.

\section{Related Work}
Recent years have witnessed remarkable advancements in the field of AIGIQA. Qu \textit{et al.} \cite{IP-IQA} develop IP-IQA, a CLIP-based framework that jointly processes AIGIs and their text prompts. Yuan \textit{et al.} \cite{yuan2023pscr} introduce PSCR, which allows for inter-comparison among images in the database. Zhou \textit{et al.} \cite{AMFF} introduce AMFF-Net, which employs multi-scale image inputs to extract features and predict multiple quality dimensions. Peng  \textit{et al.} \cite{IPCE_2024_CVPR} propose ICPE, an enhanced version of LIQE that incorporates text-image correlation analysis. Yu \textit{et al.} \cite{Yu_2024_CVPR} present SF-IQA, a unique metric combining quality and similarity assessments through score fusion. Yang \textit{et al.} \cite{MOE-AGIQA} present MoE-AGIQA, a hybrid expert system integrating visual and semantic perception. Tang \textit{et al.} \cite{tang2025clip} present CLIP-AGIQA, leveraging CLIP's multimodal knowledge for quality regression.  These methods typically employ sophisticated feature extraction strategies to explore and integrate multi-level visual information (spanning local details to global semantics). While such strategies can enhance model performance, they substantially increase computational complexity, making it difficult to meet the real-time inference requirements. In contrast,  we propose a  knowledge distillation-based optimization framework PKT in this paper,  which achieves efficient inference while ensuring effective processing of multi-level visual information.

\section{Approach}
\begin{figure*}[!t]
\centering
\includegraphics[width=0.98\textwidth]{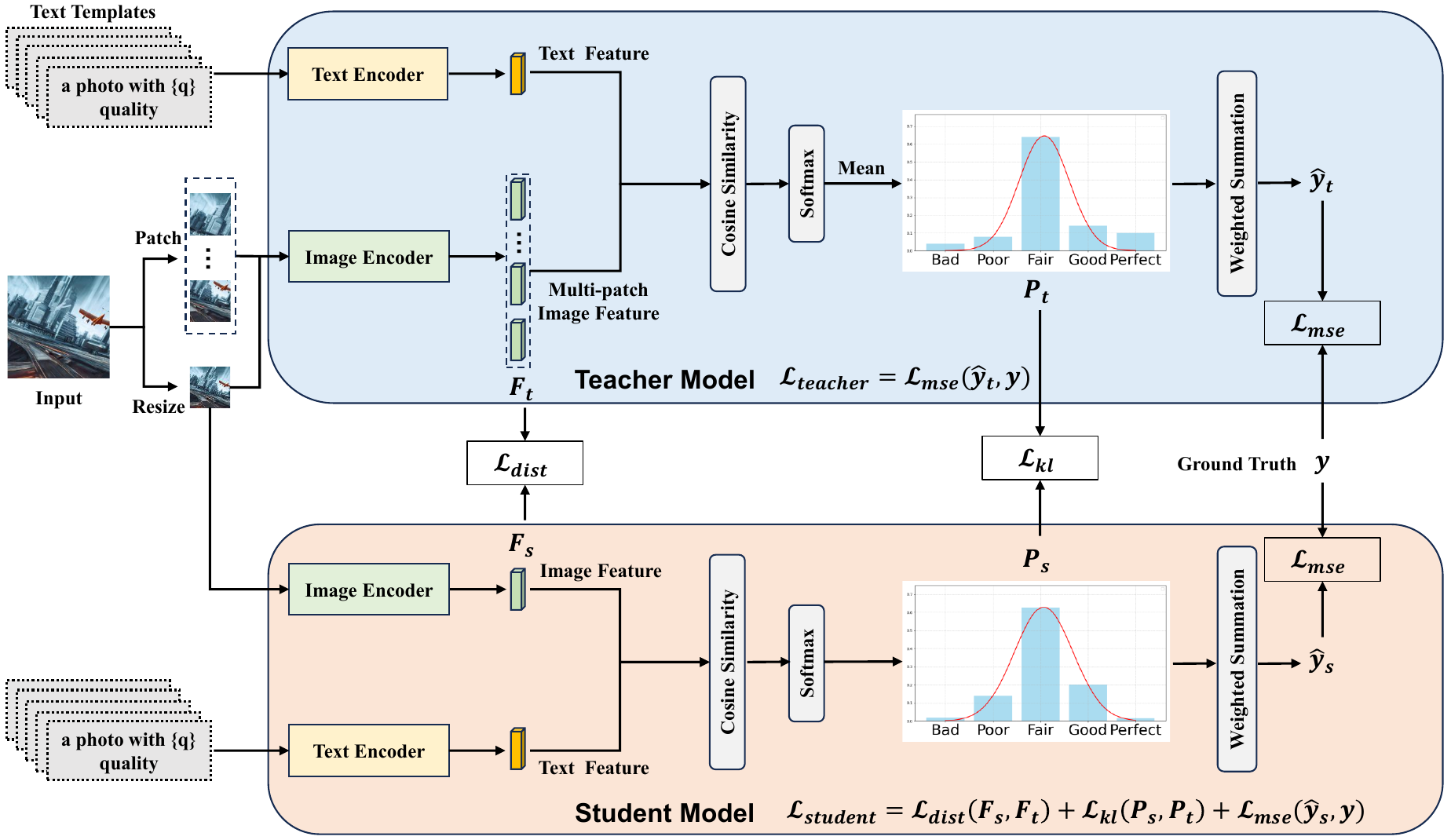}
\caption{The pipeline of the proposed PKT framework. The framework comprises two key components: (1) a teacher model employing a local-global hybrid feature processing mechanism to generate precise supervision signals, and (2) a student model relying solely on global processing that efficiently inherits the teacher's representational capacity through our designed multi-level supervision mechanism. Specifically, this supervision mechanism achieves dual knowledge transfer: deep representation migration at the feature level coupled with refined knowledge distillation at the output level.}
\label{PKT}
\end{figure*}

In this section, we first introduce the fundamental concepts and problem formulation. Subsequently, we provide a comprehensive elaboration on the operational mechanism of the proposed Patch Knowledge Transfer (PKT) approach. Fig. \ref{PKT} illustrates the overall framework of PKT.

\subsection{Pipeline Overview}

In recent years, the application of the CLIP model in image quality assessment (IQA) has attracted widespread attention \cite{CLIPIQA,LIQE,AMFF,IPCE_2024_CVPR,tang2025clip,TSP}. In this work, both the teacher and student models largely follow the architecture of LIQE. LIQE \cite{LIQE} computes cosine similarity between the image and a structured textual description: “a photo with \{q\} quality”, where q corresponds to one of five quality levels (bad, poor, fair, good, perfect) associated with five score values $V=\{v_i\}_{i=1}^5$. The similarities are transformed into probabilities via a Softmax function, and a weighted sum is applied to produce the final image quality score.

Given an image $X$, the teacher model samples $N$ image patches $X_p=\{x^i_p\}_{i=1}^N$ from the input image and inputs them along with the scaled image $X_r$ to the encoder, while the student model processes only the scaled image. This process can be represented as:
\begin{align}
    F_t = E^I_t(X_p, X_r), \label{Eq1} \\
    F_s = E^I_s(X_r), \label{Eq2} 
\end{align}
Here, $F_t=\{F^i_t\}_{i=1}^{N+1}$ and $F_s$ represent the multi-patch image features extracted by the teacher model and the single-scale image features extracted by student model, respectively. $E^I_t$ and $E^I_s$ denote the image encoders of the teacher and student models, respectively. For the designed text template, the teacher and student models use their respective text encoders to extract text features $F^T_t$ and $F^T_s$. Subsequently, we calculate the cosine similarity between the image features and the text features as follows:
\begin{align}
  S_t =   \frac{F^T_t \cdot F_t}{\lVert F^T_t \rVert  \lVert F_t \rVert}, \label{Eq3} \\
  S_s =   \frac{F^T_s \cdot F_s}{\lVert F^T_s \rVert \lVert F_s \rVert}, \label{Eq4}
\end{align}
Here, $S_t=\{S^i_t\}_{i=1}^{N+1}$ and $S_s$ represent the cosine similarity calculated by the teacher model and the student model, respectively. We then use a Softmax function to transform the cosine similarities $S_t=\{S^i_t\}_{i=1}^{N+1}$ and $S_s$ into probabilities $\mathbb{P}_t=\{\mathbb{P}^i_t\}_{i=1}^{N+1}$ and $P_s=\{p^i_s\}_{i=1}^{5}$ for the five quality levels. The output probability distribution $P_t=\{p^i_t\}_{i=1}^{5}$ of the teacher model is obtained by averaging the predicted probabilities on five quality levels for all image patches and the scaled image, which can be expressed as:
\begin{align}
  P_t =  \frac{1}{N+1} \sum_{i=1}^{N+1} \mathbb{P}^i_t, \label{Eq5} 
\end{align}
Finally, we calculate the final predicted score $\hat{y_t}$ and $\hat{y_s}$ of the teacher model and the student model through a weighted summation, which can be represented as:
\begin{align}
  \hat{y}_t =  \sum_{i=1}^{5} p^i_t \times v_i,   i=1, \cdots, 5 , \label{Eq6} \\
  \hat{y}_s =  \sum_{i=1}^{5} p^i_s \times v_i,   i=1, \cdots, 5 , \label{Eq7} 
\end{align}

\subsection{Patch Knowledge Transfer}
This paper introduces Patch Knowledge Transfer, whose key innovation lies in the design of a knowledge distillation mechanism that enables the student model to inherit visual priors extracted from multi-region features by the teacher model, thereby achieving a balance between accuracy and efficiency. The details are as follows.

\noindent
\textbf{Training Loss.} The training of the teacher model is achieved by minimizing the Mean Squared Error (MSE) loss between its predicted scores $\hat{y_t}$ and the true quality scores $y$:
\begin{align}
  \mathcal{L}_{\text{teacher}} = \mathcal{L}_{\text{mse}}(\hat{y_t}, y ) = \| \hat{y_t} - y \|_2^2 ,\label{Eq8} 
\end{align}
In order for the student model to effectively learn from the teacher model’s knowledge, we introduce multi-level supervision during the training process:
1) Feature-level knowledge transfer: By constraining the similarity between last layer features of the student model and the corresponding features of the teacher model; 2) Output-level knowledge distillation: Ensuring that the output probability distribution of the student model approaches the predicted distribution of the teacher model. To align the dimensions between the student and teacher models during feature distillation, we applied a simple average aggregation to multiple features output by the teacher model. Specifically, the overall loss function of the student model can be expressed as:  
\begin{align}
  \mathcal{L}_{\text{student}} &=   \mathcal{L}_{\text{dist}}(F_s, F_t)  + \mathcal{L}_{\text{kl}}(P_s,P_t) + \mathcal{L}_{\text{mse}}(\hat{y_s}, y) \nonumber \\ 
  &= (1 - \frac{F_s \cdot \frac{1}{N+1} \sum_{i=1}^{N+1} F^i_t }{\lVert F_s \rVert  \lVert \frac{1}{N+1} \sum_{i=1}^{N+1} F^i_t \rVert}) + \sum P_t \text{log} \frac{P_t}{P_s} \nonumber \\ &+ \| \hat{y_s} - y \|_2^2, \label{Eq9} 
\end{align}
Where $\mathcal{L}_{\text{dist}}$ is the feature-level similarity loss between the student and teacher models’ last layer outputs, $\mathcal{L}_{\text{kl}}$ is the output-level distillation loss between the predicted output probability distributions of the student and teacher models, $\mathcal{L}_{\text{mse}}$ is the mean squared error loss between the predicted quality score and the true score. 

\noindent
\textbf{One-Stage Training and Two-Stage Training.} This study explores two knowledge distillation training paradigms: 1) Joint training mode (PKT-1), where the teacher and student models are trained end-to-end with synchronized optimization of parameters, with the objective function being:
\begin{align}
  \mathcal{L}_{\text{total}} &=   \mathcal{L}_{\text{dist}}(F_s, F_t)  + \mathcal{L}_{\text{kl}}(P_s, P_t) + \mathcal{L}_{\text{mse}}(\hat{y_s}, y) \nonumber \\ &+ \mathcal{L}_{\text{mse}}(\hat{y_t}, y ); \label{Eq10} 
\end{align}
2) Stage-wise training mode (PKT-2), where the teacher model is first trained independently, and its parameters are then fixed to guide the training of the student model.

\section{Experiment}


\subsection{Databases and Evaluation Criteria}
We evaluate our approach on 4 AIGIQA benchmarks, including AGIQA-1K \cite{AGIQA1K}, AGIQA-3K \cite{AGIQA-3K}, AIGCIQA2023 \cite{aigciqa2023}, and PKU-AIGIQA-4K \cite{PKUAIGIQA4K}. This study employs two widely adopted image quality assessment (IQA) metrics: the Spearman Rank Correlation Coefficient (SRCC) to evaluate prediction monotonicity and the Pearson Linear Correlation Coefficient (PLCC) to assess prediction accuracy.

\subsection{Implementation Details}
Our experiments are conducted on the NVIDIA A100, using PyTorch 1.11.0 and CUDA 11.3 for both training and testing.  The batch size $B$ is set to $8$ for training and  set to $20$ for testing. We utilize the Adam optimizer \cite{kingma2014adam} with a learning rate of $5 \times 10^{-6}$ and weight decay of $1 \times 10^{-3}$. Unless otherwise specified, the default configurations in this study are as follows: the image encoder employs ViT-B/32 \cite{vit}, the teacher model adopts an overlapping patch sampling strategy with 9 sampled patches, and a two-stage training strategy is adopted by default during model training.  Additionally, data augmentation techniques including random flipping are implemented during training. For dataset splitting, we randomly split the dataset into training set and test set at a ratio of 4:1.

\subsection{Results and Analysis}
\noindent
\textbf{Comparisons with SOTA IQA Methods.} 
We conduct extensive comparisons between our method and 9 SOTA IQA models:  MUSIQ \cite{musiq}, HyperIQA \cite{hyperiqa}, StairIQA \cite{StairIQA}, MANIQA \cite{MANIQA}, LIQE \cite{LIQE}, Re-IQA \cite{Re-IQA}, PSCR\cite{yuan2023pscr}, AMFF-Net \cite{AMFF}, and CLIP-AGIQA \cite{tang2025clip}. The experimental results in TABLE \ref{tab1} demonstrate the relative performance across 4 AIGIQA databases, including AGIQA-1K, AGIQA-3K, AIGCIQA2023, and PKU-AIGIQA-4K. Experimental results demonstrate that our proposed PKT framework achieves a superior balance between model performance and computational efficiency compared to existing methods, as evidenced by two key aspects:

\noindent
\textbf{(1) Significant Performance Gains at Comparable Computational Cost}:
When compared to methods with similar FLOPs, PKT shows clear performance advantages. For example, on the AGIQA-1K benchmark:
\begin{itemize}
\item PKT-2 vs. HyperIQA: +0.0473 (SRCC) / +0.0275 (PLCC)
\item PKT-2 vs. StairIQA: +0.0367 (SRCC) / +0.0288 (PLCC)
\end{itemize}
\noindent
 \textbf{(2) Drastic FLOPs Reduction While Maintaining Competitive Performance}:
Compared to high-performance large models, PKT achieves comparable or better accuracy with significantly lower computational cost. For example, on the AGIQA-1K benchmark:
\begin{itemize}
\item PKT-2 vs. MANIQA: -95.98G FLOPs , +0.0203 (SRCC) / +0.0103 (PLCC)
\item PKT-2 vs. MUSIQ: -113.88G FLOPs , +0.0501 (SRCC) / +0.0337 (PLCC)
\item PKT-2 vs. LIQE: -41.25G FLOPs , +0.008 (SRCC) / +0.007 (PLCC)
\end{itemize}
Notably, our PKT-2 achieves SOTA average performance across 4 benchmark datasets. On the AGIQA-1K and AGIQA-3K databases, PKT-2 exhibits particularly dominant advantages, outperforming all competing methods comprehensively.

\noindent
\textbf{Comparisons with Multiple Patch-based Teacher and Baseline.}
To further validate the effectiveness of our proposed method, we systematically compared it with the teacher model using local-global hybrid processing and the student model without PKT (Baseline). The experiments were comprehensively verified using multiple image encoders, including ViT-B/32, ViT-B/16 and ViT-L/14 \cite{vit}. As shown in TABLE \ref{tab2}, the teacher model demonstrates superior performance due to its higher computational complexity (e.g., the ViT-L/14 teacher model has FLOPs as high as 540.75G). By introducing the PKT method, the student model not only significantly surpasses the baseline under the same computational cost but also approaches or even exceeds the teacher model's performance on multiple metrics (e.g., evaluation results of ViT-B/16 and ViT-L/14), fully demonstrating effectiveness of PKT in knowledge transfer. For example, on the PKU-AIGIQA-4K benchmark, when using ViT-B/32 as the image encoder, our student model achieves performance improvements of 0.0178 and 0.0147 in SRCC and PLCC metrics, respectively, while maintaining comparable computational costs to the baseline. More notably, the proposed method reduces computational costs by 67.7\% while achieving performance comparable to the teacher model, highlighting its advantage in balancing model efficiency and performance.

\begin{table*}[t]
\centering
\caption{Comparisons with SOTA IQA methods. Results marked with an asterisk (*) are directly taken from \cite{AMFF}, while those without an asterisk are reproduced from our experiments. The best and second best performances are bolded and \uline{underlined}, respectively.}
\resizebox{\textwidth}{!}{
\begin{tabular}{l|c|c|c|c|c|c|c|c|c|c}
\toprule
\multirow{2}{*}{Method}  & \multirow{2}{*}{FLOPs}  & \multirow{2}{*}{\#Params}  & \multicolumn{2}{c|}{AGIQA-1K} & \multicolumn{2}{c|}{AGIQA-3K} & \multicolumn{2}{c|}{AIGCIQA2023}   & \multicolumn{2}{c}{PKU-AIGIQA-4K}\\
\cmidrule(r){4-11}
 &  &  &  SRCC & PLCC & SRCC & PLCC   & SRCC & PLCC& SRCC & PLCC\\
\cmidrule(r){1-11}
MUSIQ \cite{musiq}     & 126.52G & 78.55M & 0.8506 & 0.8850&  0.8338&  0.8698&   0.8261 & 0.8382    & 0.6801&0.6697  \\
HyperIQA \cite{hyperiqa}  & 8.67G &  27.38M  & 0.8534  & 0.8912 & 0.8495 & 0.8923 & 0.8159 &  0.8212   & 0.7144& 0.7180 \\
StairIQA \cite{StairIQA} & 5.11G & 30.49M &  0.8640 & 0.8899 &  0.8543&  0.8943&  0.8313&  0.8376     & 0.7247 &0.7145\\  
MANIQA \cite{MANIQA}  & 108.62G &   128.51M &  0.8804 & 0.9084 & 0.8916 & 0.9194 & 0.8412 &  0.8540   &0.7800 & 0.7800 \\
LIQE \cite{LIQE}   & 53.89G & 84.23M  & 0.8927 &  0.9117 & 0.9009 & 0.9220 & \textbf{0.8608} &  \textbf{0.8774}  & \textbf{0.8030} & \textbf{0.8001} \\
Re-IQA* \cite{Re-IQA} &  33.09G & 40.29M & - & - & 0.8187&  0.8799 & 0.8144 & 0.8317 & - & - \\
PSCR \cite{yuan2023pscr} & 303.55G&  87.60M & 0.8902  & \uline{0.9161} & 0.8879 & 0.9204 & 0.8558 & 0.8659  &0.7895  & 0.7821 \\
AMFF-Net* \cite{AMFF} & 41.48G &  51.30M & - &  - & 0.8565 & 0.9050 &0.8409  & 0.8537 &- &-\\
CLIP-AGIQA \cite{tang2025clip} & 22.90G &  84.29M &  0.8645 &  0.8823 & 0.8939 & 0.9165 &  0.8422 &0.8501   & 0.7735 & 0.7719\\
\cmidrule(r){1-11}
PKT-1   & 12.64G & 84.23M &  \uline{0.8935} & 0.9066 &  \uline{0.9071} &  \uline{0.9242}  &  \uline{0.8560} &  \uline{0.8723} & 0.7900 & 0.7851 \\
PKT-2 & 12.64G & 84.23M & \textbf{0.9007} & \textbf{0.9187} & \textbf{0.9077} &  \textbf{0.9269} & 0.8554 & 0.8695  &   \uline{0.8005} &  \uline{0.7978}   \\
\bottomrule
\end{tabular}
}
\label{tab1}
\end{table*}

\begin{table*}[t]
\centering
\caption{Comparisons with multiple patch-based teacher and baseline. FLOPs calculation only includes the parameters actually used during inference. Best results for each backbone are highlighted in bold.}
\resizebox{\textwidth}{!}{
\begin{tabular}{l|c|c|c|c|c|c|c|c|c|c}
\toprule
\multirow{2}{*}{Backbone} & \multirow{2}{*}{Method}  & \multirow{2}{*}{FLOPs}  & \multicolumn{2}{c|}{AGIQA-1K} & \multicolumn{2}{c|}{AGIQA-3K} & \multicolumn{2}{c|}{AIGCIQA2023}   & \multicolumn{2}{c}{PKU-AIGIQA-4K}\\
\cmidrule(r){4-11}
 &  &  &  SRCC & PLCC & SRCC & PLCC   & SRCC & PLCC& SRCC & PLCC\\
 \cmidrule(r){1-11}
\multirow{3}{*}{ViT-B/32} & Baseline   & 12.64G  & 0.8884 & 0.9029 & 0.9044 & 0.9236  & 0.8495 & 0.8671 & 0.7827 & 0.7831 \\
& Teacher & 39.16G  & \textbf{0.9017} & 0.9137 & \textbf{0.9103} & \textbf{0.9292} &  \textbf{0.8612} & \textbf{0.8795}  &   \textbf{0.8069} & \textbf{0.8057}   \\
& Student & 12.64G  & 0.9007 & \textbf{0.9187} & 0.9077 &  0.9269 & 0.8554 & 0.8695  &  0.8005 & 0.7978   \\
 \cmidrule(r){1-11}
\multirow{3}{*}{ViT-B/16} & Baseline   & 20.96G  &  0.8935& 0.9025 & \textbf{0.9134} & 0.9310 & 0.8589 & 0.8731 &  0.8109 & 0.8095 \\
& Teacher & 122.39G  & 0.8917 & \textbf{0.9079} & 0.9110 & 0.9297 &\textbf{0.8628}  &  \textbf{0.8763} & \textbf{0.8236} &  \textbf{0.8156} \\
& Student & 20.96G  & \textbf{0.8953} & 0.9076 & 0.9127  & \textbf{0.9319} & 0.8599 & 0.8748 & 0.8177 & 0.8113 \\
 \cmidrule(r){1-11}
\multirow{3}{*}{ViT-L/14} & Baseline  & 73.69G & 0.8819  & 0.8949 & 0.9195 & 0.9367 & 0.8692 & 0.8802 &  0.8262&  0.8237 \\
& Teacher & 540.75G  & \textbf{0.9036}  & \textbf{0.9145}  & 0.9199 & 0.9364 &  \textbf{0.8707} & \textbf{0.8835} & 0.8271 & 0.8221 \\
& Student &  73.69G & 0.8939 & 0.9103  & \textbf{0.9233} &  \textbf{0.9395} & 0.8685 &  0.8816 & \textbf{0.8282} & \textbf{0.8269} \\
\bottomrule
\end{tabular}
}
\label{tab2}
\end{table*}

\begin{table}[t]
\centering
\caption{Study on feature alignment strategies for feature-level knowledge transfer. Best results are highlighted in bold.}
\resizebox{0.48\textwidth}{!}{
\begin{tabular}{l|c|c|c|c}
\toprule
\multirow{2}{*}{Method} & \multicolumn{2}{c|}{AGIQA-1K} & \multicolumn{2}{c}{AGIQA-3K} \\
\cmidrule(r){2-5}
 & SRCC & PLCC &  SRCC & PLCC\\
\cmidrule(r){1-5}
Manhattan Distance & 0.8956 & 0.9143 &0.9060 &  0.9237 \\
Euclidean Distance & 0.8902 & 0.9122 & 0.9072 & 0.9246  \\
Cosine Similarity  & \textbf{0.9007} & \textbf{0.9187} & \textbf{0.9077} &  \textbf{0.9269}    \\
\bottomrule
\end{tabular}}
\label{tab4}
\end{table}

\noindent
\textbf{Study on feature alignment strategies.}
To thoroughly optimize the effectiveness of feature-level knowledge distillation, we conduct a systematic comparative study on different feature alignment strategies. The experimental results are presented in TABLE \ref{tab4}. Our findings demonstrate that, compared to Manhattan and Euclidean Distance minimization approaches, the cosine similarity maximization strategy exhibits significant advantages. This observation can be explained from the perspective of feature space geometry: cosine similarity focuses on directional consistency of feature vectors rather than absolute value matching, allowing the student model to learn the teacher's feature distribution patterns more flexibly without being constrained by feature scale differences.

\noindent
\textbf{Ablation Study.}
In this section, we systematically investigate the impact of different loss function combinations on model performance, comparing seven distinct configurations. The baseline model employs only $\mathcal{L}_{\text{mse}}$ without any teacher knowledge. Ablation studies on loss functions presented in TABLE \ref{tab8} reveal three key findings: First, while $\mathcal{L}_{\text{dist}}$ alone underperforms the baseline, its combination with $\mathcal{L}_{\text{mse}}$ improves SRCC and PLCC by 0.0115 and 0.0089 respectively compared to baseline, demonstrating that feature-level transfer requires task-specific constraints for effective performance gain. Second, $\mathcal{L}_{\text{kl}}$ alone surpasses the baseline, confirming the direct efficacy of output-level distillation. Third, the combined use of all three losses achieves optimal performance, validating our multi-level distillation strategy - feature-level and output-level knowledge transfer exhibit complementary effects, whose synergy maximizes distillation benefits.

\begin{table}[t]
\centering
\caption{Ablation Study. Best results are highlighted in bold.}
\resizebox{0.48\textwidth}{!}{
\begin{tabular}{ccc|c|c|c|c}
\toprule
\multicolumn{3}{c|}{Method} & \multicolumn{2}{c|}{AGIQA-1K} & \multicolumn{2}{c}{AGIQA-3K} \\
\cmidrule(r){1-7}
 $\mathcal{L}_{\text{dist}}$ & $\mathcal{L}_{\text{kl}}$ & $\mathcal{L}_{\text{mse}}$ & SRCC & PLCC& SRCC & PLCC  \\
\cmidrule(r){1-7}
$\checkmark$ & - & - & 0.8364  & 0.8658 &0.8766 &  0.9066 \\
 - & $\checkmark$ & - & 0.8951  & 0.9174 &0.9036 &  0.9225 \\
- & - & $\checkmark$  & 0.8884 & 0.9029 & 0.9044 & 0.9236\\
 - & $\checkmark$ & $\checkmark$ & 0.8922  & 0.9128 & 0.9057 &  0.9241 \\
 $\checkmark$ & - & $\checkmark$ &  0.8999 & 0.9118 & 0.9067& 0.9252  \\
 $\checkmark$ & $\checkmark$ & - & 0.9001  & \textbf{0.9189} & 0.9054 & 0.9260  \\
 $\checkmark$ & $\checkmark$ & $\checkmark$ & \textbf{0.9007} & 0.9187 & \textbf{0.9077} &  \textbf{0.9269}   \\
\bottomrule
\end{tabular}}
\label{tab8}
\end{table}

\subsection{Visualization and Analysis}
\begin{figure}[!t]
\centering
\subfloat[Baseline]{\includegraphics[width=0.23\textwidth]{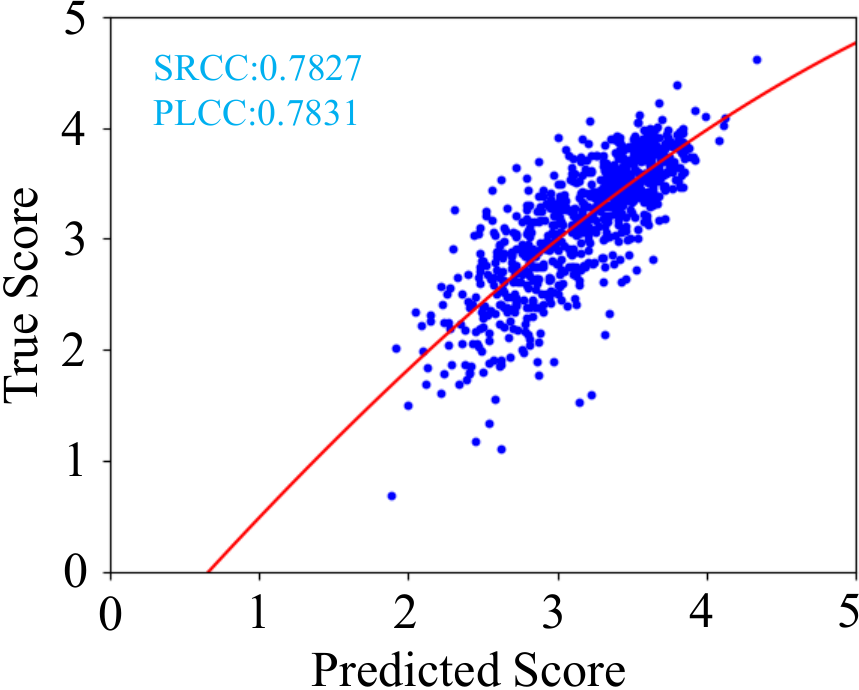}%
\label{baseline}}
\hfil
\subfloat[PKT]{\includegraphics[width=0.23\textwidth]{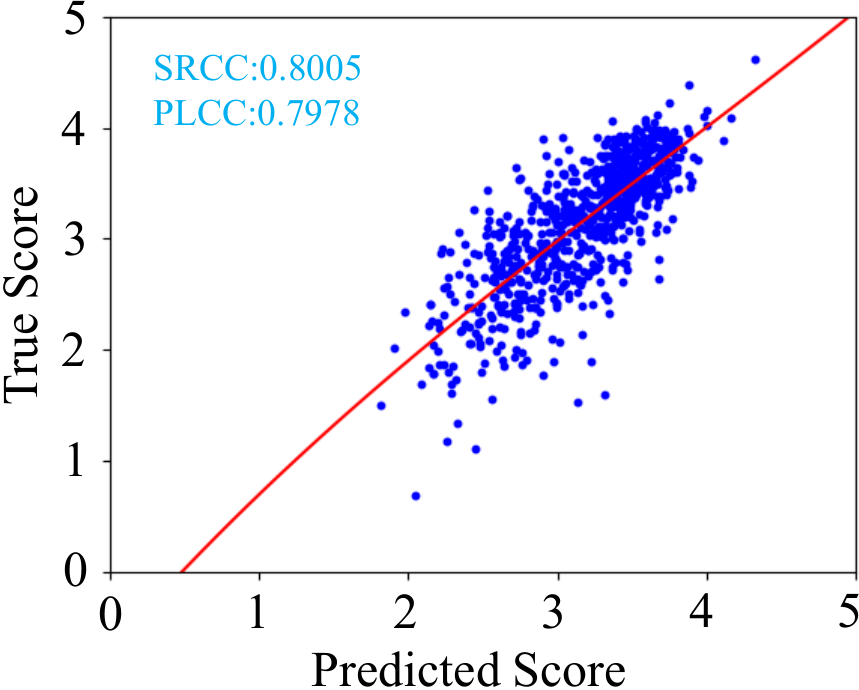}%
\label{sp}}
\caption{Visualization of score prediction scatter plots for Baseline and PKT methods on the PKU-AIGIQA-4K database. `Baseline' denotes the student model without PKT, while `PKT' denotes the student model with PKT.}
\label{fig3}
\end{figure}

To further demonstrate the effectiveness of our method, we first visualize the score prediction results of the baseline model and our proposed PKT method on the test set of the PKU-AIGIQA-4K database using scatter plots, as shown in Fig. \ref{fig3}. We observed that, compared to the baseline model, the predicted points of the proposed method are more densely distributed near the diagonal, which intuitively reflects that the student model using the PKT method achieves higher prediction accuracy. Additionally, we provide a comparison of the inference time between the student and teacher models trained using the proposed method, as shown in TABLE \ref{tab9}. Taking the teacher model as the baseline, the student model requires, on average, only 36\% of the inference time for evaluating the same number of images. This means that, while achieving comparable or even superior performance to the teacher model, the student model improves inference efficiency by 64\%. These results further validate the effectiveness of our method.

\begin{table}[t]
\centering
\caption{Comparison of inference time between the student model and the teacher model.}
\resizebox{0.48\textwidth}{!}{
\begin{tabular}{lcccc}
\toprule
Model & Batch Size &Patch& Resolution  &Inference Time \\
\cmidrule(r){1-5}
Teacher & 20 & 10 & 224 $\times$ 224  & 61.05ms \\
Student & 20 & 1 & 224 $\times$ 224  &21.98ms  \\
\bottomrule
\end{tabular}}
\label{tab9}
\end{table}

\section{Conclusion}
This paper presents PKT, a knowledge distillation-based optimization framework that addresses the critical challenge of balancing efficiency and accuracy in AIGIQA task. The proposed framework employs a teacher-student architecture with multi-level supervision during training to achieve efficient knowledge transfer. Extensive experiments on 4  AIGIQA benchmarks demonstrate that PKT outperforms existing IQA methods, maintaining high accuracy while significantly improving inference efficiency - particularly suitable for deployment scenarios with stringent real-time requirements. We expect the introduction of PKT will provide new insights and solutions for efficient quality assessment of AIGIs.

\bibliographystyle{IEEEtran}
\bibliography{icme2026references}

@String(CVPR= {IEEE Conf. Comput. Vis. Pattern Recog.})

@String(ICME = {Int. Conf. Multimedia and Expo})

@String(AAAI = {AAAI})

@String(CVPRW= {IEEE Conf. Comput. Vis. Pattern Recog. Worksh.})

@String(CVPR  = {CVPR})

@String(ICME  =	{ICME})

@String(CVPRW= {CVPRW})

@article{yuan2023pscr,
  title={PSCR: Patches Sampling-based Contrastive Regression for AIGC Image Quality Assessment},
  author={Yuan, Jiquan and Cao, Xinyan and Cao, Linjing and Lin, Jinlong and Cao, Xixin},
  journal={arXiv preprint arXiv:2312.05897},
  year={2023}
}

@article{yuan2023pkui2iqa,
  title={PKU-I2IQA: An Image-to-Image Quality Assessment Database for AI Generated Images},
  author={Yuan, Jiquan and Cao, Xinyan and Li, Changjin and Yang, Fanyi and Lin, Jinlong and Cao, Xixin},
  journal={arXiv preprint arXiv:2311.15556},
  year={2023}
}

@inproceedings{musiq,
  title={Musiq: Multi-scale image quality transformer},
  author={Ke, Junjie and Wang, Qifei and Wang, Yilin and Milanfar, Peyman and Yang, Feng},
  booktitle={Proceedings of the IEEE/CVF international conference on computer vision},
  pages={5148--5157},
  year={2021}
}

@article{MANIQA,
  title={MANIQA: Multi-dimension Attention Network for No-Reference Image Quality Assessment},
  author={Sidi Yang and Tianhe Wu and Shu Shi and Shan Gong and Ming Cao and Jiahao Wang and Yujiu Yang},
  journal={2022 IEEE/CVF Conference on Computer Vision and Pattern Recognition Workshops (CVPRW)},
  year={2022},
  pages={1190-1199},
  url={https://api.semanticscholar.org/CorpusID:248240148}
}

@inproceedings{LIQE,
  title={Blind image quality assessment via vision-language correspondence: A multitask learning perspective},
  author={Zhang, Weixia and Zhai, Guangtao and Wei, Ying and Yang, Xiaokang and Ma, Kede},
  booktitle={Proceedings of the IEEE/CVF conference on computer vision and pattern recognition},
  pages={14071--14081},
  year={2023}
}

@inproceedings{LinearityIQA,
  title={Norm-in-norm loss with faster convergence and better performance for image quality assessment},
  author={Li, Dingquan and Jiang, Tingting and Jiang, Ming},
  booktitle={Proceedings of the 28th ACM International conference on multimedia},
  pages={789--797},
  year={2020}
}

@INPROCEEDINGS{StairIQA,
  author={Sun, Wei and Duan, Huiyu and Min, Xiongkuo and Chen, Li and Zhai, Guangtao},
  booktitle={2022 IEEE International Symposium on Broadband Multimedia Systems and Broadcasting (BMSB)}, 
  title={Blind Quality Assessment for in-the-Wild Images via Hierarchical Feature Fusion Strategy}, 
  year={2022},
  volume={},
  number={},
  pages={01-06},
  doi={10.1109/BMSB55706.2022.9828590}}

@INPROCEEDINGS{hyperiqa,
  author={Su, Shaolin and Yan, Qingsen and Zhu, Yu and Zhang, Cheng and Ge, Xin and Sun, Jinqiu and Zhang, Yanning},
  booktitle={2020 IEEE/CVF Conference on Computer Vision and Pattern Recognition (CVPR)}, 
  title={Blindly Assess Image Quality in the Wild Guided by a Self-Adaptive Hyper Network}, 
  year={2020},
  volume={},
  number={},
  pages={3664-3673},
  doi={10.1109/CVPR42600.2020.00372}}

@article{AMFF,
  title={Adaptive mixed-scale feature fusion network for blind AI-generated image quality assessment},
  author={Zhou, Tianwei and Tan, Songbai and Zhou, Wei and Luo, Yu and Wang, Yuan-Gen and Yue, Guanghui},
  journal={IEEE Transactions on Broadcasting},
  year={2024},
  publisher={IEEE}
}

@InProceedings{IPCE_2024_CVPR,
    author    = {Peng, Fei and Fu, Huiyuan and Ming, Anlong and Wang, Chuanming and Ma, Huadong and He, Shuai and Dou, Zifei and Chen, Shu},
    title     = {AIGC Image Quality Assessment via Image-Prompt Correspondence},
    booktitle = {Proceedings of the IEEE/CVF Conference on Computer Vision and Pattern Recognition (CVPR) Workshops},
    month     = {June},
    year      = {2024},
    pages     = {6432-6441}
}

@article{KD,
  title={Distilling the Knowledge in a Neural Network},
  author={Geoffrey E. Hinton and Oriol Vinyals and Jeffrey Dean},
  journal={ArXiv},
  year={2015},
  volume={abs/1503.02531},
  url={https://api.semanticscholar.org/CorpusID:7200347}
}

@article{TSP,
  title={AI-Generated Image Quality Assessment Based on Task-Specific Prompt and Multi-Granularity Similarity},
  author={Jili Xia and Lihuo He and Fei Gao and Kaifan Zhang and Leida Li and Xinbo Gao},
  journal={ArXiv},
  year={2024},
  volume={abs/2411.16087},
  url={https://api.semanticscholar.org/CorpusID:274234606}
}

@InProceedings{MOE-AGIQA,
    author    = {Yang, Junfeng and Fu, Jing and Zhang, Wei and Cao, Wenzhi and Liu, Limei and Peng, Han},
    title     = {MoE-AGIQA: Mixture-of-Experts Boosted Visual Perception-Driven and Semantic-Aware Quality Assessment for AI-Generated Images},
    booktitle = {Proceedings of the IEEE/CVF Conference on Computer Vision and Pattern Recognition (CVPR) Workshops},
    month     = {June},
    year      = {2024},
    pages     = {6395-6404}
}

@inproceedings{AIGIQA20K,
  title={AIGIQA-20K: A Large Database for AI-Generated Image Quality Assessment},
  author={Chunyi Li and Tengchuan Kou and Yixuan Gao and Yu Shan Cao and Wei Sun and Zicheng Zhang and Yingjie Zhou and Zhichao Zhang and Weixia Zhang and Haoning Wu and Xiaohong Liu and Xiongkuo Min and Guangtao Zhai},
  year={2024},
  url={https://api.semanticscholar.org/CorpusID:268889245}
}

@article{AGIQA1K,
  title={A Perceptual Quality Assessment Exploration for AIGC Images},
  author={Zhang, Zicheng and Li, Chunyi and Sun, Wei and Liu, Xiaohong and Min, Xiongkuo and Zhai, Guangtao},
  journal={arXiv preprint arXiv:2303.12618},
  year={2023}
}

@ARTICLE{AGIQA-3K,
  author={Li, Chunyi and Zhang, Zicheng and Wu, Haoning and Sun, Wei and Min, Xiongkuo and Liu, Xiaohong and Zhai, Guangtao and Lin, Weisi},
  journal={IEEE Transactions on Circuits and Systems for Video Technology}, 
  title={AGIQA-3K: An Open Database for AI-Generated Image Quality Assessment}, 
  year={2023},
  pages={1-1},
  doi={10.1109/TCSVT.2023.3319020}}

@article{aigciqa2023,
  title={Aigciqa2023: A large-scale image quality assessment database for ai generated images: from the perspectives of quality, authenticity and correspondence},
  author={Wang, Jiarui and Duan, Huiyu and Liu, Jing and Chen, Shi and Min, Xiongkuo and Zhai, Guangtao},
  journal={arXiv preprint arXiv:2307.00211},
  year={2023}
}

@article{PKUAIGIQA4K,
  title={PKU-AIGIQA-4K: A Perceptual Quality Assessment Database for Both Text-to-Image and Image-to-Image AI-Generated Images},
  author={Jiquan Yuan and Fanyi Yang and Jihe Li and Xinyan Cao and Jinming Che and Jinlong Lin and Xixin Cao},
  journal={ArXiv},
  year={2024},
  volume={abs/2404.18409},
  url={https://api.semanticscholar.org/CorpusID:269449873}
}

@inproceedings{CLIPIQA,
  title={Exploring clip for assessing the look and feel of images},
  author={Wang, Jianyi and Chan, Kelvin CK and Loy, Chen Change},
  booktitle={Proceedings of the AAAI Conference on Artificial Intelligence},
  volume={37},
  number={2},
  pages={2555--2563},
  year={2023}
}

@article{kingma2014adam,
  title={Adam: A method for stochastic optimization},
  author={Kingma, Diederik P and Ba, Jimmy},
  journal={arXiv preprint arXiv:1412.6980},
  year={2014}
}

@inproceedings{IP-IQA,
  title={Bringing textual prompt to ai-generated image quality assessment},
  author={Qu, Bowen and Li, Haohui and Gao, Wei},
  booktitle={2024 IEEE International Conference on Multimedia and Expo (ICME)},
  pages={1--6},
  year={2024},
  organization={IEEE}
}

@InProceedings{Yu_2024_CVPR,
    author    = {Yu, Zihao and Guan, Fengbin and Lu, Yiting and Li, Xin and Chen, Zhibo},
    title     = {SF-IQA: Quality and Similarity Integration for AI Generated Image Quality Assessment},
    booktitle = {Proceedings of the IEEE/CVF Conference on Computer Vision and Pattern Recognition (CVPR) Workshops},
    month     = {June},
    year      = {2024},
    pages     = {6692-6701}
}

@inproceedings{tang2025clip,
  title={CLIP-AGIQA: Boosting the Performance of AI-Generated Image Quality Assessment with CLIP},
  author={Tang, Zhenchen and Wang, Zichuan and Peng, Bo and Dong, Jing},
  booktitle={International Conference on Pattern Recognition},
  pages={48--61},
  year={2025},
  organization={Springer}
}

@inproceedings{AlignIQA,
  title={Align-IQA: Aligning Image Quality Assessment Models with Diverse Human Preferences via Customizable Guidance},
  author={Junfeng Yang and Jing Fu and Zhen Zhang and Limei Liu and Qin Li and Wei Zhang and Wenzhi Cao},
  booktitle={ACM Multimedia},
  year={2024},
  url={https://api.semanticscholar.org/CorpusID:273646294}
}

@article{vit,
  title={An image is worth 16x16 words: Transformers for image recognition at scale},
  author={Dosovitskiy, Alexey and Beyer, Lucas and Kolesnikov, Alexander and Weissenborn, Dirk and Zhai, Xiaohua and Unterthiner, Thomas and Dehghani, Mostafa and Minderer, Matthias and Heigold, Georg and Gelly, Sylvain and others},
  journal={arXiv preprint arXiv:2010.11929},
  year={2020}
}

@article{Re-IQA,
  title={Re-IQA: Unsupervised Learning for Image Quality Assessment in the Wild},
  author={Avinab Saha and Sandeep Mishra and Alan Conrad Bovik},
  journal={2023 IEEE/CVF Conference on Computer Vision and Pattern Recognition (CVPR)},
  year={2023},
  pages={5846-5855},
  url={https://api.semanticscholar.org/CorpusID:257913460}
}

\appendix

\subsection{Similarity-based Region-Weighted Adaptive Feature Distillation}
The loss function used for feature-level knowledge transfer mentioned above merely guides the feature learning of the student model by simply averaging multiple regional features of the teacher model, which may exhibit certain limitations. We argue that during feature transfer, different regional features should contribute differentially. To achieve more refined feature-level distillation, we propose  Similarity-based Region-Weighted Adaptive Feature Distillation (SRWAFD). It introduces no additional learnable parameters. Instead, it directly utilizes the similarity between patch-wise features from the teacher model and the student features as weighting coefficients to differentiate the importance of distinct regional features, which dynamically evaluates the importance of each region in the teacher's features for the student's feature learning, thereby enabling more efficient feature knowledge distillation.  The implementation involves two key steps: first, computing the cosine similarity $C=\{c_i\}_{i=1}^{N+1}$ between each regional feature of the teacher and the student's features as follws:
\begin{align}
   c_i =   \frac{F_s \cdot F^i_t}{\lVert F_s \rVert  \lVert F^i_t \rVert} , i=1,  \cdots,N+1; \label{Eq.16} 
\end{align}
second, converting the cosine similarity into normalized attention weights $W=\{w_i\}_{i=1}^{N+1}$ via the Softmax function:
\begin{align}
   w_i =   \frac{e^{c_i}}{\sum_{j=1}^{N+1}e^{c_j}},   i=1, \cdots, N+1 \label{Eq.17} 
\end{align}
Ultimately, the loss function can be expressed as:
\begin{align}
    \mathcal{L}_{\text{dist}}(F_s, F_t)   
  = 1 - \frac{F_s \cdot  \sum_{i=1}^{N+1} w_i F^i_t }{\lVert F_s \rVert  \lVert \sum_{i=1}^{N+1} w_i F^i_t \rVert}  \label{Eq11} 
\end{align}

\subsection{Additional Experimental Results}
In this section, we present additional experimental results that were not included in the main text.

\noindent
\textbf{Results on AIGIQA-20K.} TABLE \ref{tab1} compares the performance of multiple patch-based teacher models against baseline methods on the AIGIQA20K\cite{AIGIQA20K} dataset.

\noindent
\textbf{Study on Different Patches Sampling method.} In this work, the teacher model adopts a local-global hybrid input approach that combines image patches with a downscaled full image for feature extraction. To thoroughly investigate the impact of different image patches sampling strategies in the teacher model on the performance of the student model, we systematically compare three typical sampling methods: random patches sampling (RPS), non-overlapping patches sampling (NOPS), and overlapping patches sampling (OPS). Experimental results presented in TABLE \ref{tab2} demonstrate that the OPS strategy exhibits the best performance stability, delivering significant and consistent improvements across all tested datasets. In contrast, while RPS may achieve optimal results on certain datasets, it suffers from high performance variability and lacks stability. NOPS performs slightly worse than OPS across all three tested databases. Based on this analysis, we ultimately select OPS as the default patches sampling strategy, ensuring robustness and providing more reliable supervision signals for knowledge transfer to the student model.

\noindent
\textbf{Study on different number of sampled patches.} 
To further optimize guidance effectiveness of the teacher model, we conduct an in-depth study on how the number of sampled image patches affects the performance of the student model. Based on the OPS strategy, we design comparative experiments to evaluate model performance with 9, 16, and 25 sampled patches.  As shown in TABLE \ref{tab3}, the results indicate that the relationship between patch count and model performance is not strictly positive. When sampling 9 patches, the student model achieves the best average performance across all evaluation metrics. This finding suggests that a moderate amount of local information provides the most discriminative supervision signals for knowledge transfer, whereas excessive patches may introduce feature redundancy, ultimately impairing generalization. Additionally, computational cost increases linearly with patch count, yet performance gains remain marginal.  Based on this analysis, we set the default number of sampled patches to 9 when employing the OPS strategy. Additionally, TABLE \ref{tab4} analyzes how the teacher model's performance varies with an increasing number of sampled patches.

\noindent
\textbf{Comparison of homogeneous self-distillation and heterogeneous distillation.} To further investigate the impact of model architecture in knowledge distillation, we systematically compared two paradigms: homogeneous self-distillation (ViT-B/32  $\rightarrow$ ViT-B/32) and heterogeneous distillation (ViT-B/16  $\rightarrow$ ViT-B/32, ViT-L/14  $\rightarrow$ ViT-B/32). When performing heterogeneous distillation where the student and teacher dimensions do not match—for instance, when distilling from ViT-L/14 to ViT-B/32, with student features of shape $(B, 512)$ and teacher features of $(B, N, 768)$—we directly apply linear interpolation to the student’s features to adjust them to $(B, 768)$ for dimension alignment. Notably, this process does not introduce any additional parameters or mapping heads. As shown in TABLE \ref{tab5}, the experimental results reveal that although ViT-B/16 and ViT-L/14 teacher models outperform ViT-B/32 individually (shown in Table 2 of the mian text), their knowledge transfer efficacy is inferior to homogeneous self-distillation—achieving lower scores on most evaluation metrics. Notably, these larger teacher models also incur substantially higher training costs. Based on this empirical analysis, we ultimately select the homogeneous self-distillation scheme as our knowledge transfer mechanism, due to its more balanced computational efficiency and performance.

\begin{table}[t]
\centering
\caption{Comparisons with multiple patch-based teacher and baseline on AIGIQA20K. `Baseline' denotes the global-processing-only student model without PKT.  Best results for each backbone are highlighted in bold.}
\begin{tabular}{l|c|c|c|c}
\toprule
\multirow{2}{*}{Backbone} & \multirow{2}{*}{Method}  & \multirow{2}{*}{FLOPs}     & \multicolumn{2}{c}{AIGIQA20K}\\
\cmidrule(r){4-5}
 &  &  &  SRCC & PLCC \\
 \cmidrule(r){1-5}
\multirow{3}{*}{ViT-B/32} & Baseline   & 12.64G  & 0.8474 &0.8899 \\
& Teacher & 39.16G  & \textbf{0.8732} & \textbf{0.9005}  \\
& Student & 12.64G   &0.8575  & 0.8956   \\
\bottomrule
\end{tabular}
\label{tab1}
\end{table}

\begin{table}[t]
\centering
\caption{Study on different patches sampling method. Best results are highlighted in bold.}
\resizebox{0.48\textwidth}{!}{
\begin{tabular}{l|c|c|c|c|c|c|c|c}
\toprule
\multirow{2}{*}{Method} & \multicolumn{2}{c|}{AGIQA-1K} & \multicolumn{2}{c|}{AGIQA-3K} & \multicolumn{2}{c|}{PKU-AIGIQA-4K}  & \multicolumn{2}{c}{AIGIQA-20K} \\
\cmidrule(r){2-9}
 & SRCC & PLCC &  SRCC & PLCC &  SRCC & PLCC &  SRCC & PLCC\\
\cmidrule(r){1-9}
RPS & 0.8924 & 0.9078 &0.9056 & 0.9243 & \textbf{0.8026} & \textbf{0.7988}  & 0.8521 & 0.8913 \\
NOPS  & 0.9000 & 0.9135 & \textbf{0.9088} & 0.9264 &   0.7966 & 0.7912 & 0.8554 & 0.8948 \\
OPS  & \textbf{0.9007} & \textbf{0.9187} & 0.9077 &  \textbf{0.9269}  &  0.8005 & 0.7978   & \textbf{0.8575}  & \textbf{0.8956} \\
\bottomrule
\end{tabular}}
\label{tab2}
\end{table}

\begin{table}[t]
\centering
\caption{Study on different number of sampled patches. Best results are highlighted in bold.}
\resizebox{0.48\textwidth}{!}{
\begin{tabular}{l|c|c|c|c|c|c}
\toprule
\multirow{2}{*}{Number} & \multicolumn{2}{c|}{AGIQA-1K} & \multicolumn{2}{c|}{AGIQA-3K} & \multicolumn{2}{c}{PKU-AIGIQA-4K} \\
\cmidrule(r){2-7}
 & SRCC & PLCC &  SRCC & PLCC &  SRCC & PLCC\\
\cmidrule(r){1-7}
9 & \textbf{0.9007} & \textbf{0.9187} & \textbf{0.9077} &  \textbf{0.9269}  &  0.8005 & 0.7978   \\
16  & 0.8936 & 0.9098 & 0.9066 & 0.9268 &0.8054 & 0.7989 \\
25  & 0.8993 & 0.9114 &0.9063  & 0.9253 & \textbf{0.8145} &  \textbf{0.8070} \\
\bottomrule
\end{tabular}}
\label{tab3}
\end{table}

\begin{table}[t]
\centering
\caption{Study on different number of sampled patches.}
\resizebox{0.48\textwidth}{!}{
\begin{tabular}{l|c|c|c|c|c|c|c}
\toprule
\multirow{2}{*}{Number} & \multirow{2}{*}{Model} & \multicolumn{2}{c|}{AGIQA-1K} & \multicolumn{2}{c|}{AGIQA-3K} & \multicolumn{2}{c}{PKU-AIGIQA-4K} \\
\cmidrule(r){3-8}
& & SRCC & PLCC &  SRCC & PLCC &  SRCC & PLCC\\
 \cmidrule(r){1-8}
\multirow{2}{*}{9} & teacher & 0.9017 & 0.9137 & 0.9103 & 0.9292  &   0.8069 & 0.8057\\
&  student & 0.9007 & 0.9187 & 0.9077 &  0.9269  &  0.8005 & 0.7978   \\
\cmidrule(r){1-8}
\multirow{2}{*}{16} & teacher & 0.8931  & 0.9086 & 0.9029 & 0.9246 &0.8132 & 0.8106 \\
&  student & 0.8936 & 0.9098 & 0.9066 & 0.9268 &0.8054 & 0.7989 \\
\cmidrule(r){1-8}
\multirow{2}{*}{25} & teacher & 0.8967 & 0.9090 & 0.9062 & 0.9253  & 0.8098  & 0.8062 \\
&  student & 0.8993 & 0.9114 &0.9063  & 0.9253 & 0.8145&  0.8070 \\

\bottomrule
\end{tabular}}
\label{tab4}
\end{table}

\begin{table}[t]
\centering
\caption{Comparison of homogeneous self-distillation and heterogeneous distillation. Best results are highlighted in bold.}
\resizebox{0.48\textwidth}{!}{
\begin{tabular}{c|c|c|c|c|c|c|c}
\toprule
\multirow{2}{*}{Student} & \multirow{2}{*}{Teacher} & \multicolumn{2}{c|}{AGIQA-1K} & \multicolumn{2}{c|}{AGIQA-3K}  & \multicolumn{2}{c}{PKU-AIGIQA-4K} \\
\cmidrule(r){3-8}
 &   &  SRCC & PLCC &  SRCC & PLCC&  SRCC & PLCC\\
\cmidrule(r){1-8}
\multirow{3}{*}{ViT-B/32} & ViT-L/14 & 0.9045 & 0.9139 & \textbf{0.9079} & 0.9258  & 0.7989 & 0.7949 \\
& ViT-B/16 & \textbf{0.9066} & 0.9162 & 0.9043 & 0.9233 & 0.7964 & 0.7925\\
&  ViT-B/32 & 0.9007 & \textbf{0.9187} & 0.9077 &  \textbf{0.9269}  & \textbf{0.8005}  & \textbf{0.7978}\\
\bottomrule
\end{tabular}}
\label{tab5}
\end{table}

\begin{table}[t]
\centering
\caption{Cross-Dataset Evaluation. The notation ``3K $\rightarrow$ 1K'' indicates training on the AGIQA-3K dataset and testing on the AGIQA-1K dataset; ``1K $\rightarrow$ 3K'' indicates training on the AGIQA-1K dataset and testing on the AGIQA-3K dataset. }
\resizebox{0.48\textwidth}{!}{
\begin{tabular}{l|c|c|c|c}
\toprule
\multirow{2}{*}{method} & \multicolumn{2}{c|}{3K $\rightarrow$ 1K} & \multicolumn{2}{c}{1K $\rightarrow$ 3K}\\
\cmidrule(r){2-5}
& SRCC & PLCC & SRCC & PLCC \\
\cmidrule(r){1-5}
LinearityIQA \cite{LinearityIQA}  &0.6368 &  0.6828 & 0.5742&0.6023\\
MUSIQ \cite{musiq}    &0.7000 & 0.7720 & 0.5548 &0.5591 \\
HyperIQA \cite{hyperiqa} & 0.7047 &    0.7742   & 0.5725&   0.6159  \\
StairIQA \cite{StairIQA}    &0.6911 &  0.7814 & 0.5706& 0.5898\\  
MANIQA \cite{MANIQA}  &0.6666 & 0.7577 & 0.5669& 0.5625\\
LIQE \cite{LIQE} & 0.7202 &  0.7995 & 0.7485& 0.7605\\
MoE-AGIQA-v2 \cite{MOE-AGIQA} &0.7101 &  0.8008 &0.7483 & 0.7506\\
CLIP-AGIQA\cite{tang2025clip}  & 0.7238 & 0.7964 &  0.5457 &  0.5612\\
\cmidrule(r){1-5}
PKT-1  & 0.7409 & 0.8202 & 0.7547 &0.7655 \\
PKT-2  &  \textbf{0.7619} &  \textbf{0.8276} &  \textbf{0.7640} &  \textbf{0.7794} \\
\bottomrule
\end{tabular}}
\label{cross}
\end{table}

\noindent
\textbf{Additional experimental results about study on feature aggregation strategies.}  We conduct experiments to compare the effects of different feature aggregation strategies on student model performance, including: (1) simple average; (2) learned weights, which assigns weights to different regions via a fully connected layer and a sigmoid activation function; and (3) the proposed  Similarity-based Region-Weighted Adaptive Feature Distillation (SRWAFD). As shown in TABLE \ref{tab6}, the results indicate that learned weights underperforms even simple average, whereas the proposed SRWAFD method further improves model performance without introducing any additional parameters, demonstrating its superiority.

\noindent
\textbf{Additional experimental results about comparisons under different temperature settings.} Temperature $T$ is a key parameter in knowledge distillation, designed to soften probability distributions during training. We further investigated the impact of different temperature coefficients on student model performance. Experimental results in TABLE \ref{tab7} indicate that distillation achieves the best performance when no temperature scaling is applied (i.e., $T$=1) on the AGIQA-1K and AGIQA-3K dataset. Nevertheless, tuning $T$ may yield benefits, as shown in the supplementary appendix. In this work, we set $T$=1 by default without employing temperature scaling to soften the outputs of the student and teacher models. This is because temperature softening does not always lead to a performance gain.  

\noindent
\textbf{Impact of different loss weights.} In the main text, the loss function for the students of PKT-2 is as follows:
\begin{align}
  \mathcal{L}_{\text{student}} =   \mathcal{L}_{\text{dist}}(F_s, F_t)  + \mathcal{L}_{\text{kl}}(P_s,P_t) + \mathcal{L}_{\text{mse}}(\hat{y_s}, y), \label{Eq9} 
\end{align}
In the above formulation, the coefficients for different weights are set to 1. Here, we investigate the impact of assigning different weights to distinct loss functions. The new loss function is defined as follows:
\begin{align}
  \mathcal{L}_{\text{student}} =  \alpha \cdot \mathcal{L}_{\text{dist}}(F_s, F_t)  + \beta \cdot \mathcal{L}_{\text{kl}}(P_s,P_t) + \gamma \cdot \mathcal{L}_{\text{mse}}(\hat{y_s}, y), \label{Eq10} 
\end{align}
Experimental results in Table \ref{taba} show that different loss weights have minimal impact on model performance.

\noindent
\textbf{Additional experimental results about robustness experiments.} To validate the robustness of the proposed method, we conducted experiments under four different dataset splits using random seeds 0, 1, 2, and 42. The complete experimental results are presented in TABLE \ref{tab8}, TABLE \ref{tab9}, TABLE \ref{tab10}, TABLE \ref{tab11}, TABLE \ref{tab12}, TABLE \ref{tab13}, and TABLE \ref{tab14}. Additionally, we conduct cross-dataset evaluations as shown in TABLE \ref{cross}. These results further demonstrate the robustness of the proposed method

\subsection{Visualization and Analysis}
This section presents additional visualization results to further demonstrate our method's effectiveness. 

\begin{figure*}[!t]
\centering
\subfloat[Baseline]{\includegraphics[width=0.48\textwidth]{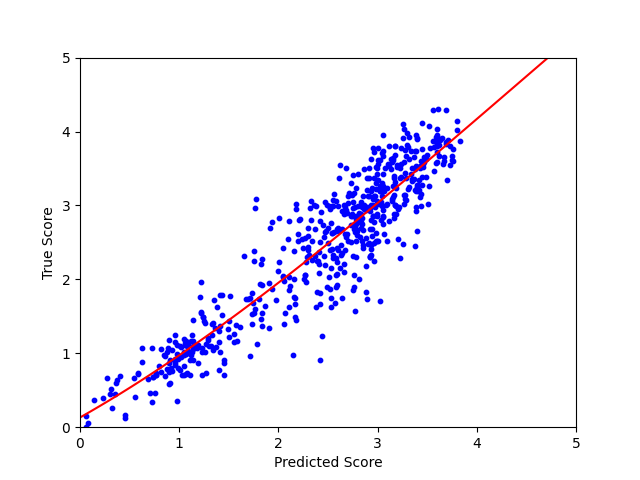}%
\label{baseline}}
\hfil
\subfloat[PKT]{\includegraphics[width=0.48\textwidth]{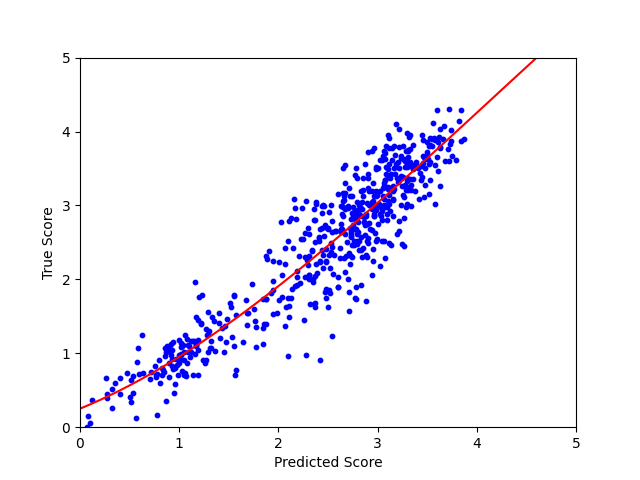}%
\label{sp}}
\caption{Visualization of score prediction scatter plots for Baseline and PKT methods on the AGIQA-3K database. `Baseline' denotes the student model without PKT, while `PKT' denotes the student model with PKT.}
\label{2}
\end{figure*}

Fig. \ref{2} visualizes the score prediction scatter plots comparing the Baseline and PKT methods on the AGIQA-3K database. The denser distribution of data points along the diagonal in our PKT method demonstrates its enhanced prediction accuracy compared to the baseline approach. This visual evidence confirms that our knowledge transfer framework effectively improves score prediction reliability.

\subsection{Discussion}
\noindent
\textbf{A short discussion about situations where the student model occasionally outperforms the teacher model.} We believe this is because the student model does not merely fit the labels of the dataset but learns from the richer signals of the teacher model, which encourages the formation of smoother and more generalized feature representations. It helps avoid overfitting to the details that the teacher model might have already overfitted.

\begin{table*}[t]
\centering
\caption{Comparison of different feature aggregation strategy. The best performances are bolded.}
\resizebox{\textwidth}{!}{
\begin{tabular}{lccccccccccc}
\toprule
\multicolumn{2}{c}{\multirow{2}{*}{Method}}    & \multicolumn{2}{c}{AGIQA-1K} & \multicolumn{2}{c}{AGIQA-3K} & \multicolumn{2}{c}{AIGCIQA2023}   & \multicolumn{2}{c}{PKU-AIGIQA-4K} & \multicolumn{2}{c}{Average}\\
\cmidrule(r){3-12}
 &   &   SRCC & PLCC & SRCC & PLCC   & SRCC & PLCC& SRCC & PLCC& SRCC & PLCC\\
 \cmidrule(r){1-12}
\multirow{2}{*}{PKT-1}  &Average &  \textbf{0.8935} & 0.9066 &  \textbf{0.9071} &  \textbf{0.9242}  &  \textbf{0.8560} &  \textbf{0.8723} & 0.7900 & 0.7851 &0.8617 & 0.8721\\
& SRWAKD &  0.8925 & \textbf{0.9088} & 0.9065  & 0.9241   & 0.8552  &  \textbf{0.8723} & \textbf{0.7939} & \textbf{0.7931}  & \textbf{0.8620} & \textbf{0.8746} \\
\cmidrule(r){1-12}

\multirow{2}{*}{PKT-2} & Average & 0.9007 & \textbf{0.9187} & 0.9077 &  \textbf{0.9269} & 0.8554 & 0.8695  &   0.8005 &  0.7978   & 0.8661& 0.8782\\
& SRWAKD & \textbf{0.9023}  & 0.9156 & \textbf{0.9080}  & 0.9266   &  \textbf{0.8559} & \textbf{0.8704}  & \textbf{0.8040} & \textbf{0.8013}  & \textbf{0.8676} & \textbf{0.8785}\\
\bottomrule
\end{tabular}
}
\label{tab6}
\end{table*}

\begin{table*}[t]
\centering
\caption{Performance comparison of student models under different temperature settings. Best results are highlighted in bold.}
\resizebox{\textwidth}{!}{
\begin{tabular}{l|c|c|c|c|c|c|c|c|c}
\toprule
\multirow{2}{*}{Method} & \multirow{2}{*}{temp}   & \multicolumn{2}{c|}{AGIQA-1K} & \multicolumn{2}{c|}{AGIQA-3K} & \multicolumn{2}{c|}{AIGCIQA2023}   & \multicolumn{2}{c}{PKU-AIGIQA-4K}\\
\cmidrule(r){3-10}
 &  &   SRCC & PLCC & SRCC & PLCC   & SRCC & PLCC& SRCC & PLCC\\
  \cmidrule(r){1-10}
\multirow{5}{*}{PKT-1} & 1  &  0.8935 & 0.9066 &  0.9071 &  0.9242  &  0.8560 & 0.8723 & 0.7900 & 0.7851 \\
& 2   & 0.8947&0.9117 & 0.9079& 0.9250&0.8527 & 0.8671&\textbf{0.7982} & \textbf{0.7971}\\
& 3    & 0.8938&0.9094 & 0.9075 & 0.9242&\textbf{0.8577} & \textbf{0.8727}& 0.7954& 0.7923\\
& 4   & \textbf{0.8963} & \textbf{0.9119}  &0.9085 & 0.9249&0.8541 &0.8687 & 0.7980 &0.7941 \\
& 5    &0.8944 &0.9086 & \textbf{0.9104} & \textbf{0.9264} & 0.8551&0.8698&0.7958 & 0.7942\\
 \cmidrule(r){1-10}
\multirow{5}{*}{PKT-2} & 1     & 0.9007 & \textbf{0.9187} & 0.9077 &  \textbf{0.9269} & 0.8554 & 0.8695  &  0.8005 &  0.7978  \\
& 2    &0.9011 & 0.9152&\textbf{0.9078} &0.9256 & 0.8544& 0.8706& \textbf{0.8049} & \textbf{0.8001} \\
& 3     & 0.9003& 0.9146& 0.9067&0.9253 & 0.8553 & 0.8704& 0.8031& 0.7981\\
& 4    &0.9005 &0.9164 & 0.9069& 0.9238& \textbf{0.8578} & \textbf{0.8707} & 0.8037& 0.7995\\
& 5    & \textbf{0.9012}& 0.9123&0.9072 &0.9254 &0.8542  &0.8703  &0.8047 &0.7988 \\
\bottomrule
\end{tabular}
}
\label{tab7}
\end{table*}

\begin{table*}[t]
\centering
\caption{Impact of different loss weights.  Best results are highlighted in bold.}
\resizebox{\textwidth}{!}{
\begin{tabular}{l|c|c|c|c|c|c|c|c|c|c|c}
\toprule
\multirow{2}{*}{Method} & \multirow{2}{*}{$\alpha$} & \multirow{2}{*}{$\beta$} & \multirow{2}{*}{$\gamma$}  & \multicolumn{2}{c|}{AGIQA-1K} & \multicolumn{2}{c|}{AGIQA-3K} & \multicolumn{2}{c|}{AIGCIQA2023}   & \multicolumn{2}{c}{PKU-AIGIQA-4K}\\
\cmidrule(r){5-12}
 &  &  & &  SRCC & PLCC & SRCC & PLCC   & SRCC & PLCC& SRCC & PLCC\\
 \cmidrule(r){1-12}
\multirow{5}{*}{PKT-2} & 1   &1 &  1& 0.9007 & \textbf{0.9187} & 0.9077 &  \textbf{0.9269} & 0.8554 & 0.8695  &  0.8005 &  0.7978  \\
& 0.6  &0.2 & 0.2 & 0.9027 & 0.9147& 0.9057 &0.9253  & \textbf{0.8559}  & 0.8704& 0.8011 & 0.7972\\
& 0.5 & 0.25 & 0.25 & \textbf{0.9030} & 0.9142 &0.9062 & 0.9250& 0.8552& 0.8699& 0.8039 & \textbf{0.7996} \\
& 0.4  & 0.3 & 0.3 & 0.9005 & 0.9146 & 0.9070 &0.9257 & 0.8546 & 0.8670 & \textbf{0.8047} & 0.7986\\
& 0.2  & 0.4 &  0.4 & 0.9007 & 0.9164 & \textbf{0.9101} &0.9257  & 0.8558 & \textbf{0.8705} &0.8002 & 0.7968\\
\bottomrule
\end{tabular}
}
\label{taba}
\end{table*}

\begin{table*}[t]
\centering
\caption{Comparisons with multiple patch-based teacher and baseline under 4 random dataset splits on 4 AIGIQA databases . `Baseline' denotes the global-processing-only student model without PKT.  Best results are highlighted in bold.}
\resizebox{\textwidth}{!}{
\begin{tabular}{l|c|c|c|c|c|c|c|c|c|c}
\toprule
\multirow{2}{*}{Seed} & \multirow{2}{*}{Method}  & \multirow{2}{*}{FLOPs}  & \multicolumn{2}{c|}{AGIQA-1K} & \multicolumn{2}{c|}{AGIQA-3K} & \multicolumn{2}{c|}{AIGCIQA2023}   & \multicolumn{2}{c}{PKU-AIGIQA-4K}\\
\cmidrule(r){4-11}
 &  &  &  SRCC & PLCC & SRCC & PLCC   & SRCC & PLCC& SRCC & PLCC\\
 \cmidrule(r){1-11}
\multirow{3}{*}{0} & Baseline   & 12.64G  &  0.8659 & \textbf{0.8940} & 0.8916 &  0.9190 & 0.8620 & 0.8811 & 0.8080 & 0.8146 \\
& Teacher & 39.16G  & 0.8662 & 0.8886 & \textbf{0.8921} & 0.9195  & \textbf{0.8726} & \textbf{0.8887} & \textbf{0.8334} & \textbf{0.8328}  \\
& Student & 12.64G  & \textbf{0.8689} & 0.8904 & 0.8919 &  \textbf{0.9198} & 0.8643 & 0.8848 & 0.8200 & 0.8211 \\
 \cmidrule(r){1-11}
\multirow{3}{*}{1} & Baseline   & 12.64G  & 0.8884 & 0.9029 & 0.9044 & 0.9236  & 0.8495 & 0.8671 & 0.7827 & 0.7831 \\
& Teacher & 39.16G  & \textbf{0.9017} & 0.9137 & \textbf{0.9103} & \textbf{0.9292} &  \textbf{0.8612} & \textbf{0.8795}  &   \textbf{0.8069} & \textbf{0.8057}   \\
& Student & 12.64G  & 0.9007 & \textbf{0.9187} & 0.9077 &  0.9269 & 0.8554 & 0.8695  &  0.8005 & 0.7978   \\
 \cmidrule(r){1-11}
\multirow{3}{*}{2} & Baseline   & 12.64G   & 0.8853 & 0.8965 &0.8705  & 0.9128 & 0.8648 &0.8763& 0.8055 & 0.8066\\
& Teacher & 39.16G   &0.8858 &0.8943 &0.8684 & 0.9110& \textbf{0.8771}& \textbf{0.8901}& \textbf{0.8248} & \textbf{0.8216} \\
& Student & 12.64G   & \textbf{0.8890} & \textbf{0.9054} &\textbf{0.8746} & \textbf{0.9142}&0.8660 & 0.8817& 0.8186 & 0.8154\\
 \cmidrule(r){1-11}
\multirow{3}{*}{42} & Baseline   & 12.64G  & 0.8740 & 0.9125 & 0.8787 & 0.9162  & 0.8501 &  0.8726&  0.7917 & 0.8025 \\
& Teacher & 39.16G  & 0.8749 & 0.9011 & \textbf{0.8879} &  \textbf{0.9171} & \textbf{0.8696} & \textbf{0.8859} & \textbf{0.8159} &  \textbf{0.8192} \\
& Student & 12.64G  &\textbf{ 0.8762} & \textbf{0.9171} & 0.8805 &  0.9169 & 0.8633 & 0.8803 & 0.8060 & 0.8124 \\
\bottomrule
\end{tabular}
}
\label{tab8}
\end{table*}

\begin{table*}[t]
\centering
\caption{Comparisons with SOTA IQA methods on 4 AIGIQA databases, with a random seed of 0 used for dataset splitting. The best performances are bolded.}
\resizebox{\textwidth}{!}{
\begin{tabular}{l|c|c|c|c|c|c|c|c|c|c}
\toprule
\multirow{2}{*}{Method}  & \multirow{2}{*}{FLOPs}  & \multirow{2}{*}{\#Params}  & \multicolumn{2}{c|}{AGIQA-1K} & \multicolumn{2}{c|}{AGIQA-3K} & \multicolumn{2}{c|}{AIGCIQA2023}   & \multicolumn{2}{c}{PKU-AIGIQA-4K}\\
\cmidrule(r){4-11}
 &  &  &  SRCC & PLCC & SRCC & PLCC   & SRCC & PLCC& SRCC & PLCC\\
\cmidrule(r){1-11}
LinearityIQA \cite{LinearityIQA}   & 1.82G & 12.53M   &0.8201 &0.8443 & 0.7963& 0.8167& 0.8223&0.8284 & 0.6948&0.6492 \\
MUSIQ \cite{musiq}     & 126.52G & 78.55M  & 0.8367& 0.8770& 0.8168& 0.8696 &0.8474 &0.8575 & 0.6830& 0.6868\\
HyperIQA \cite{hyperiqa}  & 8.67G &  27.38M  &0.8325 & 0.8780&0.8451 & 0.8947& 0.8327&0.8440 & 0.7539& 0.7609\\
StairIQA \cite{StairIQA} & 5.11G & 30.49M  &0.8410 &0.8768 &0.8494 & 0.9014&0.8446 &0.8525 & 0.7505& 0.7581\\ 
MANIQA \cite{MANIQA}  & 108.62G &   128.51M  & 0.8414& 0.8750& 0.8882& 0.9222& 0.5398& 0.5287 &0.8112 & 0.8096 \\
LIQE \cite{LIQE}   & 53.89G & 84.23M   &0.8579 &0.8840 &0.8964 &0.9202 & \textbf{0.8708}& \textbf{0.8925}& \textbf{0.8280}&\textbf{0.8288} \\
CLIP-AGIQA\cite{tang2025clip} & 22.90G &  84.29M  & 0.8462& 0.8579& 0.8830&0.9164 & 0.8642& 0.8731& 0.7988 & 0.8047\\
\cmidrule(r){1-11}
PKT-1   & 12.64G & 84.23M & \textbf{0.8744}&\textbf{0.9015} & \textbf{0.8990} &\textbf{0.9241} & 0.8688& 0.8862& 0.8151& 0.8209\\
PKT-2 & 12.64G & 84.23M  & 0.8689 & 0.8904 & 0.8919 &  0.9198 & 0.8643 & 0.8848 &0.8200 & 0.8211 \\
\bottomrule
\end{tabular}
}
\label{tab9}
\end{table*}

\begin{table*}[t]
\centering
\caption{Comparisons with SOTA IQA methods on 4 AIGIQA databases with a random seed of 1 used for dataset splitting. The best performances are bolded. }
\resizebox{\textwidth}{!}{
\begin{tabular}{l|c|c|c|c|c|c|c|c|c|c}
\toprule
\multirow{2}{*}{Method}  & \multirow{2}{*}{FLOPs}  & \multirow{2}{*}{\#Params}  & \multicolumn{2}{c|}{AGIQA-1K} & \multicolumn{2}{c|}{AGIQA-3K} & \multicolumn{2}{c|}{AIGCIQA2023}   & \multicolumn{2}{c}{PKU-AIGIQA-4K}\\
\cmidrule(r){4-11}
 &  &  &  SRCC & PLCC & SRCC & PLCC   & SRCC & PLCC& SRCC & PLCC\\
\cmidrule(r){1-11}
LinearityIQA \cite{LinearityIQA}   & 1.82G & 12.53M   &  0.8200 & 0.8578 & 0.8189 &  0.8309& 0.7947 &  0.8057    & 0.6493&0.6056  \\
MUSIQ \cite{musiq}     & 126.52G & 78.55M & 0.8506 & 0.8850&  0.8338&  0.8698&   0.8261 & 0.8382    & 0.6801&0.6697  \\
HyperIQA \cite{hyperiqa}  & 8.67G &  27.38M  & 0.8534  & 0.8912 & 0.8495 & 0.8923 & 0.8159 &  0.8212   & 0.7144& 0.7180 \\
StairIQA \cite{StairIQA} & 5.11G & 30.49M &  0.8640 & 0.8899 &  0.8543&  0.8943&  0.8313&  0.8376     & 0.7247 &0.7145\\  
MANIQA \cite{MANIQA}  & 108.62G &   128.51M &  0.8804 & 0.9084 & 0.8916 & 0.9194 & 0.8412 &  0.8540   &0.7800 & 0.7800 \\
LIQE \cite{LIQE}   & 53.89G & 84.23M  & 0.8927 &  0.9117 & 0.9009 & 0.9220 & \textbf{0.8608} &  \textbf{0.8774}  & \textbf{0.8030} & \textbf{0.8001} \\
CLIP-AGIQA\cite{tang2025clip} & 22.90G &  84.29M &  0.8645 &  0.8823 & 0.8939 & 0.9165 &  0.8422 &0.8501   & 0.7735 & 0.7719\\
\cmidrule(r){1-11}
PKT-1   & 12.64G & 84.23M &  0.8935 & 0.9066 &  0.9071 &  0.9242  &  0.8560 &  0.8723 & 0.7900 & 0.7851 \\
PKT-2 & 12.64G & 84.23M & \textbf{0.9007} & \textbf{0.9187} & \textbf{0.9077} &  \textbf{0.9269} & 0.8554 & 0.8695  &   0.8005 &  0.7978  \\
\bottomrule
\end{tabular}
}
\label{tab10}
\end{table*}

\begin{table*}[t]
\centering
\caption{Comparisons with SOTA IQA methods on 4 AIGIQA databases, with a random seed of 2 used for dataset splitting. The best and second best performances are bolded and \uline{underlined}, respectively.}
\resizebox{\textwidth}{!}{
\begin{tabular}{l|c|c|c|c|c|c|c|c|c|c}
\toprule
\multirow{2}{*}{Method}  & \multirow{2}{*}{FLOPs}  & \multirow{2}{*}{\#Params}  & \multicolumn{2}{c|}{AGIQA-1K} & \multicolumn{2}{c|}{AGIQA-3K} & \multicolumn{2}{c|}{AIGCIQA2023}   & \multicolumn{2}{c}{PKU-AIGIQA-4K}\\
\cmidrule(r){4-11}
 &  &  &  SRCC & PLCC & SRCC & PLCC   & SRCC & PLCC& SRCC & PLCC\\
\cmidrule(r){1-11}
LinearityIQA \cite{LinearityIQA}   & 1.82G & 12.53M   & 0.8220&0.8352 & 0.7865& 0.8085& 0.7927&0.8026 & 0.6633& 0.6307\\
MUSIQ \cite{musiq}     & 126.52G & 78.55M  & 0.8383&0.8706 &0.7954 & 0.8558 & 0.8466& 0.8515&0.6376 &0.6362 \\
HyperIQA \cite{hyperiqa}  & 8.67G &  27.38M  & 0.8486& 0.8726& 0.8273& 0.8841& 0.8299& 0.8350& 0.7396& 0.7361\\
StairIQA \cite{StairIQA} & 5.11G & 30.49M  &0.8532 &0.8784 &0.8337 & 0.8891& 0.8473& 0.8555& 0.7314& 0.7389\\ 
MANIQA \cite{MANIQA}  & 108.62G &   128.51M  & 0.8613& 0.8842&0.8645 & 0.9131&0.7174 &0.7125& 0.7456& 0.7564\\
LIQE \cite{LIQE}   & 53.89G & 84.23M   &0.8702 & 0.8937&0.8684 &0.9109 &0.8644 & 0.8816& \textbf{0.8211} & \textbf{0.8162} \\
CLIP-AGIQA\cite{tang2025clip} & 22.90G &  84.29M  & 0.8518 & 0.8565 &0.8595 &0.9069 &0.8574  &0.8662&0.8070 & 0.8036\\
\cmidrule(r){1-11}
PKT-1   & 12.64G & 84.23M & 0.8849 & 0.8993& \textbf{0.8765} & \textbf{0.9163}  & 0.8656& \textbf{0.8825} & 0.8110& 0.8125\\
PKT-2 & 12.64G & 84.23M  &\textbf{0.8890} &\textbf{0.9054} &0.8746 & 0.9142&\textbf{0.8660} & 0.8817& 0.8186 & 0.8154\\
\bottomrule
\end{tabular}
}
\label{tab11}
\end{table*}

\begin{table*}[t]
\centering
\caption{Comparisons with SOTA IQA methods on 4 AIGIQA databases, with a random seed of 42 used for dataset splitting. The best and second best performances are bolded and \uline{underlined}, respectively.}
\resizebox{\textwidth}{!}{
\begin{tabular}{l|c|c|c|c|c|c|c|c|c|c}
\toprule
\multirow{2}{*}{Method}  & \multirow{2}{*}{FLOPs}  & \multirow{2}{*}{\#Params}  & \multicolumn{2}{c|}{AGIQA-1K} & \multicolumn{2}{c|}{AGIQA-3K} & \multicolumn{2}{c|}{AIGCIQA2023}   & \multicolumn{2}{c}{PKU-AIGIQA-4K}\\
\cmidrule(r){4-11}
 &  &  &  SRCC & PLCC & SRCC & PLCC   & SRCC & PLCC& SRCC & PLCC\\
\cmidrule(r){1-11}
LinearityIQA \cite{LinearityIQA}   & 1.82G & 12.53M   &0.8124 & 0.8543& 0.8155& 0.8204& 0.7801&0.7865& 0.6742&0.6557 \\
MUSIQ \cite{musiq}     & 126.52G & 78.55M  & 0.7975& 0.8617 & 0.8294& 0.8817& 0.8161& 0.8314& 0.6980& 0.6937\\
HyperIQA \cite{hyperiqa}  & 8.67G &  27.38M  & 0.8313& 0.8786& 0.8507& 0.8987&0.8215 &0.8267 &0.7290 & 0.7325\\
StairIQA \cite{StairIQA} & 5.11G & 30.49M  &0.8493 &0.8952 &0.8510 & 0.8968&0.8269&0.8396 & 0.7357&0.7498\\ 
MANIQA \cite{MANIQA}  & 108.62G &   128.51M  &0.8509 & 0.9008&0.7821 &0.8491 & 0.8459& 0.8625&0.7952 & 0.8049\\
LIQE \cite{LIQE}   & 53.89G & 84.23M   & 0.8572& 0.8948& \textbf{0.8859}& 0.9167&0.8631 & \textbf{0.8834} & \textbf{0.8091}& 0.8112\\
CLIP-AGIQA\cite{tang2025clip} & 22.90G &  84.29M  & 0.8303 & 0.8596 & 0.8807 &0.9112 &0.8520 &0.8658 &0.7820 & 0.7944\\
\cmidrule(r){1-11}
PKT-1   & 12.64G & 84.23M &0.8742 &0.9113 & 0.8805& \textbf{0.9174}&0.8607 &0.8779 & 0.8003& 0.8093\\
PKT-2 & 12.64G & 84.23M  & \textbf{0.8762} & \textbf{0.9171} & 0.8805 &  0.9169 & \textbf{0.8633} & 0.8803 & 0.8060 & \textbf{0.8124} \\
\bottomrule
\end{tabular}
}
\label{tab12}
\end{table*}

\begin{table*}[t]
\centering
\caption{Comparisons with SOTA IQA methods under 4 random dataset splits on AGIQA-1K and AGIQA-3K database. The best performances are bolded. }
\resizebox{\textwidth}{!}{
\begin{tabular}{l|c|c|c|c}
\toprule
\multirow{2}{*}{Method} & \multicolumn{2}{c|}{AGIQA-1K} & \multicolumn{2}{c}{AGIQA-3K} \\
\cmidrule(r){2-5}
 &    SRCC & PLCC & SRCC & PLCC   \\
\cmidrule(r){1-5}
LinearityIQA \cite{LinearityIQA}     & 0.8186 $\pm$ 0.0037 & 0.8479 $\pm$ 0.0088 & 0.8043 $\pm$ 0.0134 & 0.8191 $\pm$ 0.0080 \\
MUSIQ \cite{musiq}       & 0.8308 $\pm$ 0.0200 & 0.8736 $\pm$ 0.0085 & 0.8188 $\pm$ 0.0149 & 0.8692 $\pm$ 0.0092 \\
HyperIQA \cite{hyperiqa}     & 0.8415 $\pm$ 0.0097 & 0.8801 $\pm$ 0.0068 & 0.8432 $\pm$ 0.0094 & 0.8924 $\pm$ 0.0053 \\
StairIQA \cite{StairIQA}    & 0.8519 $\pm$ 0.0083 & 0.8851 $\pm$ 0.0077 & 0.8471 $\pm$ 0.0079 & 0.8954 $\pm$ 0.0044 \\
MANIQA \cite{MANIQA}   & 0.8585 $\pm$ 0.0145 & 0.8921 $\pm$ 0.0132 & 0.8566 $\pm$ 0.0443 & 0.9010 $\pm$ 0.0301 \\
LIQE \cite{LIQE}     & 0.8695 $\pm$ 0.0144 & 0.8961 $\pm$ 0.0100 & 0.8879 $\pm$ 0.0125 & 0.9174 $\pm$ 0.0042 \\
CLIP-AGIQA\cite{tang2025clip}   & 0.8482 $\pm$ 0.0123 & 0.8641 $\pm$ 0.0106 & 0.8793 $\pm$ 0.0125 & 0.9127 $\pm$ 0.0040 \\
\cmidrule(r){1-5}
PKT-1    & 0.8818 $\pm$ 0.0080 & 0.9047 $\pm$ 0.0047 & \textbf{0.8908 $\pm$ 0.0127} & \textbf{0.9205 $\pm$ 0.0037} \\
PKT-2   & \textbf{0.8837 $\pm$ 0.0122} & \textbf{0.9079 $\pm$ 0.0113} & 0.8887 $\pm$ 0.0126 & 0.9194 $\pm$ 0.0047 \\
\bottomrule
\end{tabular}
}
\label{tab13}
\end{table*}

\begin{table*}[t]
\centering
\caption{Comparisons with SOTA IQA methods under 4 random dataset splits on AIGCIQA2023 and PKU-AIGIQA-4K  database. The best performances are bolded .}
\resizebox{\textwidth}{!}{
\begin{tabular}{l|c|c|c|c}
\toprule
\multirow{2}{*}{Method} & \multicolumn{2}{c|}{AIGCIQA2023}   & \multicolumn{2}{c}{PKU-AIGIQA-4K}\\
\cmidrule(r){2-5}
 &    SRCC & PLCC & SRCC & PLCC   \\
\cmidrule(r){1-5}
LinearityIQA \cite{LinearityIQA}     & 0.7974 $\pm$ 0.0154 & 0.8058 $\pm$ 0.0149 & 0.6704 $\pm$ 0.0166 & 0.6353 $\pm$ 0.0194 \\
MUSIQ \cite{musiq}       & 0.8341 $\pm$ 0.0134 & 0.8447 $\pm$ 0.0104 & 0.6747 $\pm$ 0.0225 & 0.6716 $\pm$ 0.0222 \\
HyperIQA \cite{hyperiqa}     & 0.8250 $\pm$ 0.0067 & 0.8317 $\pm$ 0.0086 & 0.7342 $\pm$ 0.0145 & 0.7369 $\pm$ 0.0154 \\
StairIQA \cite{StairIQA}    & 0.8375 $\pm$ 0.0086 & 0.8463 $\pm$ 0.0078 & 0.7356 $\pm$ 0.0095 & 0.7403 $\pm$ 0.0164 \\
MANIQA \cite{MANIQA}   & 0.7361 $\pm$ 0.1245 & 0.7394 $\pm$ 0.1355 & 0.7830 $\pm$ 0.0242 & 0.7877 $\pm$ 0.0213 \\
LIQE \cite{LIQE}     & \textbf{0.8648 $\pm$ 0.0037} & \textbf{0.8837 $\pm$ 0.0055} & \textbf{0.8153 $\pm$ 0.0098} & \textbf{0.8141 $\pm$ 0.0103} \\
CLIP-AGIQA\cite{tang2025clip}   & 0.8539 $\pm$ 0.0080 & 0.8638 $\pm$ 0.0084 & 0.7903 $\pm$ 0.0133 & 0.7936 $\pm$ 0.0132 \\
\cmidrule(r){1-5}
PKT-1    & 0.8628 $\pm$ 0.0049 & 0.8797 $\pm$ 0.0052 & 0.8041 $\pm$ 0.0098 & 0.8069 $\pm$ 0.0133 \\
PKT-2   & 0.8622 $\pm$ 0.0041 & 0.8791 $\pm$ 0.0058 & 0.8113 $\pm$ 0.0083 & 0.8117 $\pm$ 0.0086 \\
\bottomrule
\end{tabular}
}
\label{tab14}
\end{table*}

\end{document}